# MORPHOLOGICAL EVALUATION OF SUBWORDS VOCABULARY USED BY BETO LANGUAGE MODEL

ÓSCAR GARCÍA-SIERRA[1]

*Universidad Complutense de Madrid; dezzai*

ANA FERNÁNDEZ-PAMPILLÓN CESTEROS[2]

*Universidad Complutense de Madrid*

MIGUEL ORTEGA-MARTÍN[3]

*Universidad Complutense de Madrid; dezzai*

*Abstract*

Subword tokenization algorithms used by Large Language Models are significantly more efficient and can independently build the necessary vocabulary of words and subwords without human intervention. However, those subwords do not always align with real morphemes, potentially impacting the models' performance—though it remains uncertain when this might occur. In previous research, we proposed a method to assess the morphological quality of vocabularies, focusing on the overlap between these vocabularies and the morphemes of a given language. Our evaluation method was built on three quality measures—relevance, cohesion, and morphological accuracy—and a procedure for their assessment. By applying this method to vocabularies created by three subword tokenization algorithms—*BPE*, *Wordpiece*, and *Unigram*—we concluded that these vocabularies generally exhibit very low morphological quality. In this article, we apply this evaluation to the tokenizer of BETO, a BERT language model trained on large Spanish corpora. This evaluation, along with our previous results, helped us conclude that its vocabulary has a low morphological quality, and we also found that training the tokenizer in a larger corpus does not improve the morphological quality of the generated vocabulary. Additionally, this evaluation helps clarify the algorithm used by the tokenizer, that is, *Wordpiece*, given the inconsistencies between the authors' claims and the model's configuration.

[1] oscarg02@ucm.es; https://orcid.org/0000-0002-8828-7338
[2] apampi@filol.ucm.es; https://orcid.org/0000-0002-6606-0159
[3] m.ortega@ucm.es; https://orcid.org/0000-0002-1880-5048

1. INTRODUCTION

Tokenization is the process of splitting a text into smaller units called tokens (Friedman, 2023), which can be morphemes, words, punctuation marks, characters, among others. Tokenization is essential for processing text collections to create datasets, which are used to train artificial neural networks, as these data represent the discrete units of information from which language models are built (Friedman, 2023). In the training of current *Transformer* neural networks (Vaswani et al., 2017) to generate language models, vocabularies of tokens, which can be words and subwords, are used. A subword is a part of a word that appears frequently in the training corpora but does not necessarily correspond to a morpheme. For instance, BERT (Devlin et al., 2019) and its Spanish version, BETO (Cañete et al., 2023), use vocabularies of approximately thirty-one thousand tokens.

Among the most widely used tokenization algorithms for generating the vocabularies used by language models, three stand out: *Wordpiece* (Schuster, 2012; Wu, 2016), used by BERT; Byte-Pair Encoding (*BPE*) (Sennrich, 2015), used by the GPT family (Radford et al., 2018) and RoBERTa (Liu et al., 2019); and *Unigram*, used by Albert (Lan et al., 2019). The algorithms used to generate these vocabularies are based on the frequency with which certain characters or strings appear in the training corpora, relying on purely statistical criteria. As a result, frequently occurring words in the corpus may not be split and appear as a single token in the vocabulary, while less frequent words may be split into several higher-frequency subwords. For example, BETO's tokenizer does not segment the word *población* (population) — even though it is formed by the root *pobla* and the morpheme ##ción — and treats it as a single token. However, it segments the word *perduración* (perdurance) into the tokens *perd* and ##uración, and the word *oscilación* (oscillation) into *oscila* and ##ción.

Compared to symbolic tokenizers based on linguistic rules, statistical tokenizers have the advantage of generating vocabularies with tens of thousands of tokens from large text corpora without human supervision, which is particularly useful in the development of multilingual models. However, as several studies (Church, 2020; Bostrom and Durrett, 2020; Hoffman et al., 2021; Park, 2020) have shown, they have the drawback that subword tokenization does not largely correspond to the morphemes of the language, resulting in a loss of linguistic knowledge on which language models are based and potentially reducing the effectiveness and reliability of these models. For example, in English, Hoffman et al. (2021) found that a BERT model with a morpheme-based vocabulary outperformed another BERT model with a vocabulary generated by a *BPE* tokenizer in sentiment analysis and topic prediction tasks. Jabbar (2024), also in English, compared a GPT model with *BPE* to a GPT model with a morphological tokenizer called *MorphPiece*, showing that the latter outperformed the former on a wide variety of NLP tasks.

In previous work, we proposed a solution to the problem of how to assess morphological quality based on the results of earlier studies (García-Sierra et al., 2024). We defined the morphological quality of a language model's vocabulary as its degree of similarity to a real morpheme vocabulary of the same language and established three criteria to measure it: relevance, coherence, and morphological accuracy. Each of these criteria measures an aspect of the similarity between the vocabularies: (i) morphological relevance refers to the proportion of real language morphemes contained in the LLM vocabulary; (ii) morphological coherence measures the consistency with that words which share a morpheme also share a token that corresponds to that morpheme; and, (iii), morphological accuracy measures the tokenizer's ability to use actual language morphemes (and not other subwords) to correctly segment words in the evaluation datasets.

Additionally, we proposed a method for measuring each criterion and applied it to the evaluation of the three subword algorithms—*BPE*, *Wordpiece*, and *Unigram*—using Spanish texts, a language with greater morphological complexity than English. The results indicated that, for Spanish, all three algorithms exhibit very low morphological quality and that increasing the size of the vocabularies does not improve their morphological quality. The results also revealed some intrinsic types of errors in the tokenizers. For instance, *Unigram* and *Wordpiece* tend not to segment words, unlike *BPE*; however, *BPE* generates vocabularies with more missing morphemes than *Unigram* and *Wordpiece*, resulting in poorer tokenization, especially concerning relevance and morphological accuracy.

In this paper, we present the results of applying our evaluation model and method to the real case of the vocabulary generated by the openly available the tokenizer of BETO language model[4]. BETO is a BERT-type language model for Spanish, created in 2019 and based on the *Transformer* encoder architecture, trained exclusively with Spanish texts. Specifically, the training corpus consists of three hundred million sentences, compared to the six hundred thousand sentences used to train the tokenizers evaluated in our previous work.

The evaluation of BETO's vocabulary quality is of interest because, (i), it is on the most downloaded Spanish language model on the *Hugging Face* platform, (ii), it will allow us to determine whether its vocabulary could be a potential limitation on the model's performance and, if so, explore whether improving the vocabulary leads to a significant improvement in the model, and, (iii), it will allow us to verify whether the results obtained in García-Sierra et al. (2024) regarding the quality of vocabularies generated by *Wordpiece* and *BPE* tokenization algorithms are confirmed. More specifically, we aim to verify if the generated vocabularies exhibit low quality, if the types of errors found in each tokenizer are replicated, and if training the tokenizer with a larger corpus results in a higher morphological quality of the tokenizer's vocabulary.

The paper is organized as follows: in the second section we define the objectives of the work; in the third section we indicate the possible algorithm that have been used to generate BETO's vocabulary; in the fourth section we summarize the procedure for evaluating morphological quality; in the fifth section we describe the application of this procedure to the evaluation of the quality of the vocabulary generated by BETO's tokenizer; in the sixth section we present and discuss the results, first, the quantitative values obtained for each evaluation criterion and, later, through an error analysis. In the seventh section we discuss the algorithm used by BETO's tokenizer. In the last section

---

[4] https://github.com/dccuchile/beto

we present a summary, conclusions and future work.

## 2. OBJECTIVES OF THE STUDY

The main objective of this study is to evaluate the quality of the vocabulary used by the Spanish language model BETO, with the aim of determining:

(i) whether the findings from our previous research regarding the low quality of vocabularies generated by the *BPE* and *Wordpiece* tokenization algorithms are confirmed,

(ii) whether there is a threshold beyond which the size of the training corpus becomes independent of the vocabulary quality produced by these tokenization algorithms,

(iii) the types of errors present in the vocabularies generated by these tokenization algorithms, and

(iv) whether, based on the evaluation of BETO's vocabulary quality, it is possible to identify the specific tokenization algorithm used to generate it.

## 3. BETO'S TOKENIZER

The tokenizer used by BETO's language model has been trained on a corpus of 300 million sentences, which is five hundred times larger than the corpus used to train the tokenizers in García-Sierra et al. (2024). However, with the information currently available, it is unclear what type of tokenizer BETO uses. According to its authors, the tokenizer is based on the *BPE* algorithm (Cañete et al., 2023:3). Yet, this information conflicts with what is specified on the *Hugging Face*[5] download page and in the tokenizer's configuration file[6], where it is indicated that the tokenizer used is of the *Wordpiece* type (see Figure 1).

---

[5] https://huggingface.co/dccuchile/bert-base-spanish-wwm-uncased
[6] https://huggingface.co/dccuchile/bert-base-spanish-wwm-uncased/raw/main/tokenizer.json

```
"decoder": {
  "type": "WordPiece",
  "prefix": "##",
  "cleanup": true
},
"model": {
  "type": "WordPiece",
  "unk_token": "[UNK]",
  "continuing_subword_prefix": "##",
  "max_input_chars_per_word": 100,
  "vocab": {
```

**Figure 1**. BETO's tokenizer configuration on the Hugging Face Platform.

Our intuition at this point is that *Wordpiece* is likely the tokenizer used in the current version of BETO, not only because of the indications in the configuration file, but also because if it were *BPE*, there should be a "merges.txt" file containing the learned character merges ranked by frequency from the training process. This file is absent from BETO's repository on *Hugging Face*[7]. However, considering the authors' claims, it is possible that the vocabulary was indeed created using *BPE*, and the configuration file refers to a different type of tokenizer for the inference phase. This would impact the model's performance since different subwords would be used.

Our hypothesis is that evaluating BETO's vocabulary will help resolve this inconsistency and clarify which algorithm was actually used by the tokenizer. We will do this by comparing the quality metrics and error types with the results from our previous work.

## 4. EVALUATION METHOD OF THE MORPHOLOGICAL QUALITY IN LANGUAGE MODELS VOCABULARIES

Our proposal is that the method for assessing morphological quality should be based on measuring the similarity between the vocabulary produced by a tokenizer and the actual morpheme vocabulary of the target language model (García-Sierra et al., 2024). We proposed that the degree of similarity depends on three quality criteria, as outlined in Section 1 of the introduction: morphological relevance, coherence, and accuracy. Specifically, (i) morphological relevance refers to the proportion of actual morphemes from the language that are learned by the tokenizer; (ii) morphological coherence measures the consistency with which words which share a morpheme also share a token which corresponds to that morpheme; and (iii) morphological accuracy assesses the tokenizer's ability to use the learned morphemes to correctly segment words.

---

[7] https://huggingface.co/dccuchile/bert-base-spanish-wwm-uncased/tree/main

To evaluate these three criteria, it is necessary to: (1) construct a corresponding evaluation dataset for each criterion, and (2) calculate the degree of compliance for each. (1) For the construction of the evaluation datasets:

(i) The dataset for assessing morphological relevance should consist of a list of the language's morphemes, ideally categorized by typology (e.g. prefix, suffix, and root). Table 1 provides a description of our Spanish dataset.

| Morpheme type | Total | Example |
|---|---|---|
| Prefixes | 61 | ['des', 're', ...] |
| Sufixes | 175 | ['mos', 'ción', ...] |
| Stems | 5,000 | ['dej', 'alt', ...] |

**Table 1.** Morphological relevance evaluation dataset for Spanish (source: García Sierra et al., 2024)

(ii) The dataset for the morphological consistency criterion should consist of a list of word-morpheme pairs that includes all morphemes in the language. This list can be generated from a previous dataset of morphemes and a list of words in the language that is as comprehensive as possible. Table 2 describes our Spanish dataset.

| Morpheme type | Total | Example |
|---|---|---|
| Prefixes | 205.792 | [(*remodelación*, *re-*), …] |
| Stems | 234.564 | [(*ubicásemos*, *ubic*), …] |
| Sufixes | 1.575.207 | [(*fijación*, *-ción*), …] |
| Clitics | 3.920 | [(*dilo*, *lo*), …] |

**Table 2**. Morphological relevance evaluation dataset for Spanish (source: García Sierra et al., 2024)

(iii) The dataset for evaluating morphological accuracy should consist of a list of words, each labeled with its corresponding morphological segmentation(s). To construct this dataset, a random sample of words can be selected, ensuring that all morphemes in the language are represented at least once. Additionally, the proportional representation of grammatical categories in the dictionary of the language should be maintained. Table 3 provides a description of the dataset we constructed for Spanish.

| Part-of-Speech tag | Total | Examples |
|---|---|---|
| Nouns | 481 | ['camiones': {'NOUN': ['camion', 'es']}] |
| Verbs | 445 | ['dejan': {'VERB': ['dej', 'an']}] |
| Adjectives | 299 | ['adecuadas': {'ADJ': ['adecu', 'ad', 'a', 's']}] |
| Pronouns | 61 | ['estos': {'PRON': ['est', 'o', 's']}] |
| Adverbs | 59 | ['aquí': {'ADV': ['aquí']}] |
| Determiners | 30 | ['mis': {'DET': ['mi', 's']}] |
| Prepositions | 22 | ['desde': {'ADP': ['desde']}] |
| Conjunctions | 14 | ['pero': {'CONJ': ['pero']}] |

**Table 3**. Dataset for the evaluation of morphological accuracy (source: García Sierra et al., 2024)

(2) The following metrics will be used to calculate the level of compliance with each criterion:

    i. For morphological relevance, the traditional metrics of precision, coverage, and F1-score will be used to measure the overlap between the vocabulary tokens of the language model and the morphemes collected in the morphological relevance dataset (e.g., Table 1).

    ii. For morphological coherence, three measures will be calculated for each type of morpheme (prefix, suffix, stems, clitics):

        - Measure 1 ("segmentation into a single token"): This is the percentage of words in the coherence evaluation dataset, containing the morpheme type, that are nonetheless segmented as a single token (i.e., not segmented). This measure helps identify a type of error known as "under-segmentation."[8]

        - Measure 2 ("the morpheme is not recognized"): This is the percentage of words in the coherence evaluation dataset, containing the morpheme type and segmented into multiple tokens (subwords), where the morpheme does not correspond to any of these tokens. This measure detects errors such as "over-segmentation" or "non-recognition of the morpheme."

        - Measure 3 ("morpheme is recognized"): This is the percentage of words in the coherence evaluation dataset, containing the morpheme, that are segmented into multiple tokens, where one of the tokens corresponds to the morpheme. This measure assesses the ability of the tokenizer to recognize that type of morpheme and thus the effectiveness of the model's vocabulary in recognizing that morpheme type.

    iii. For morphological accuracy, the proportion of words is calculated in such a way that all segmentation tokens match the labeled morphemes in the accuracy evaluation dataset. The accuracy score ranges from 0 to 1 (or 0% to 100%), where 0 indicates that no words are correctly segmented and 1 (or 100%) indicates that all words are correctly segmented.

---

[8] Sobre la tipología de errores de segmentación se puede consultar la página 124 de García-Sierra et al. (2024)

# 5. EVALUATION OF THE MORPHOLOGICAL QUALITY OF BETO'S VOCABULARY

To evaluate the quality of the vocabulary in the public version of BETO available on *Hugging Face*, we applied the model and evaluation method described in the previous section. As indicated in Section 3, the tokenizer type used by BETO, according to its configuration file, is the *Wordpiece* tokenizer, although the authors of BETO specify that it employs *BPE*.

The three evaluation datasets used were those constructed in our previous work (García-Sierra et al., 2024). Details of the construction procedure can be found in that publication. Both the datasets and the algorithms used to calculate the metrics are publicly available[9].

Additionally, the quality results of the current BETO's vocabulary are presented in comparison to the quality results of two vocabularies generated in our previous work using the *Wordpiece* and *BPE* segmentation algorithms. These vocabularies consist of 31,000 tokens and were trained on the *Oscar* corpus[10], comprising 600,000 sentences, available on *Hugging Face*. For clarity, we will refer to these tokenizers as *Wordpiece*_31 and *BPE*_31, respectively. These two vocabularies serve as reference baselines against which we compare the evaluation results of BETO's vocabulary. This comparison allows us to investigate, among other aspects, whether the use of larger datasets for training tokenizers improves the morphological quality of the vocabularies, or if any of the reference vocabularies exhibit similar quality to that of BETO.

*5.1. Results and discussion*

*5.1.1. Morphological relevance*

Table 4 presents the results of BETO's vocabulary quality evaluation and the in comparison to the reference vocabularies with respect to the morphological relevance criterion. Similar to the reference vocabularies, BETO's vocabulary shows very low precision, indicating that the statistical performance of its segmentation algorithm is insufficient for adequately learning the morphological units of Spanish. The highest values are obtained in terms of coverage: BETO's vocabulary achieves 90% for prefixes, nearly 73% for suffixes, and 23% for stems. This suggests that the vocabulary includes a very high proportion of prefixes, a considerable proportion of suffixes, but a very low proportion of Spanish stems. The criteria of coherence and accuracy will determine whether these learned prefixes and suffixes are effectively used by the tokenizer when BETO is employed in inference tasks. The F1 score is low due to it being the harmonic mean of precision and coverage, and the low precision significantly impacts this value.

---

[9] https://github.com/ogarciasierra/spanish-subwords-evaluation
[10] https://huggingface.co/datasets/nthngdy/oscar-small

| Type | Totals | Tokenizer | Precision (%) | Recall (%) | F1 (%) |
| --- | --- | --- | --- | --- | --- |
| Prefixes | 61 | BETO | 0,18 | 90,16 | 0,37 |
| | | *Wordpiece*_31 | 0,17 | 88,52 | 0,35 |
| | | *BPE*_31 | 0,18 | 90,16 | 0,37 |
| Sufixes | 175 | BETO | 0,41 | 72,99 | 0,81 |
| | | *Wordpiece*_31 | 0,41 | 73,56 | 0,82 |
| | | *BPE*_31 | 0,41 | 72,41 | 0,81 |
| Stems | 5,000 | BETO | 3,84 | 23,78 | 6,61 |
| | | *Wordpiece*_31 | 8,59 | 9,97 | 9,23 |
| | | *BPE*_31 | 7,75 | 8,99 | 8,32 |

**Table 4.** Results of the evaluation of the morphological relevance of BETO's tokenizer and comparison with those of *Wordpiece* and *BPE*.

Regarding the comparison of BETO's vocabulary with the *Wordpiece*_31 and *BPE*_31 reference vocabularies, we find that:

(1) For prefixes and suffixes, the values of the three metrics are practically identical between BETO's and *BPE*_31 vocabularies, and on average, very similar across all three vocabularies. This may suggest that, at least with respect to prefixes and suffixes, the tokenizer used in BETO could be of the *BPE* type. However,

(2) there is a difference of more than ten percentage points in root morphemes between the coverage values of BETO's vocabulary (24%) and the reference vocabularies (10% and 9%). This finding suggests that training a model's tokenizer with a larger corpus improves the coverage of stems, but not that of prefixes and suffixes.

When comparing BETO's segmentation algorithm with the *Wordpiece* and *BPE* algorithms, we observe that the results between BETO and *BPE* are identical. However, the results between *Wordpiece* and *BPE* are so similar that, based solely on this criterion, we cannot definitively determine which of the two algorithms is used by BETO's tokenizer.

*5.1.2. Morphological coherence*

Table 5 shows the results of the vocabulary use of BETO's tokenizer and the reference vocabularies for the evaluation of the morphological coherence criterion.

| Morpheme type | Vocabulary | Total | Wrongly segmented as on token (%) | More than one token – correct morpheme (%) | More than one token – incorrect morpheme (%) |
|---|---|---|---|---|---|
| Prefijos | BETO | 205.792 | 0.87 | 12.20 | 86.93 |
| | *Wordpiece_31* | | 0,71 | 13,45 | 85,84 |
| | *BPE_31* | | 0,04 | 65,86 | 34,10 |
| Raíces | BETO | 234.584 | 1.95 | 17.27 | 80.78 |
| | *Wordpiece_31* | | 1,64 | 16,03 | 82,33 |
| | *BPE_31* | | 0,40 | 4,66 | 94,94 |
| Sufijos | BETO | 1.575.207 | 1.01 | 13.96 | 85.03 |

|  | *Wordpiece*_31 |  | 0,74 | 15,17 | 84,09 |
|---|---|---|---|---|---|
|  | *BPE*_31 |  | 0,11 | 10,08 | 89,81 |
| Clíticos | BETO | 3.291 | 11.43 | 53.94 | 33.63 |
|  | *Wordpiece*_31 |  | 5,61 | 61,39 | 33,00 |
|  | *BPE*_31 |  | 0,69 | 48,58 | 50,73 |
| Totales | BETO | 2.019.504 | 1,12 | 14,25 | 84,63 |
|  | *Wordpiece*_31 |  | 0,86 | 15,18 | 83,96 |
|  | *BPE*_31 |  | 0,14 | 15,21 | 84,66 |

**Table 5.** Results of the morphological coherence evaluation

As observed, BETO's tokenizer exhibits relatively low coherence values: it consistently uses the same prefixes to segment words only 12% of the time, 17% for stems, 13% for suffixes, and this figure increases to nearly 54% for clitics.

In comparison with the reference vocabularies, except for clitics, the coherence results of BETO are, unlike what was observed for relevance, similar to those of *Wordpiece*_31 and significantly different from *BPE*. Additionally, BETO's vocabulary coherence results are slightly worse than those of *Wordpiece*_31, except in the case of stems, suggesting that training the tokenizer with a larger corpus does not necessarily lead to an improvement in morphological coherence.

5.1.3. Morphological accuracy

Table 6 presents the results of the evaluation of BETO's vocabulary with respect to the morphological accuracy criterion, alongside the results of the two reference vocabularies.

| **Tokenizer** | **Accuracy (%)** | **Mean of tokens per word** |
|---|---|---|
| BETO | 14,54 | 1,71 |
| *Wordpiece*_31 | 14,54 | 1,75 |
| *BPE*_31 | 8,69 | 2,39 |

**Table 6.** Morphological Accuracy Results, Average Token Usage per Word, and Comparison with *Wordpiece* and *BPE* Results

The accuracy values for the vocabulary generated by BETO's tokenizer are quite low at 14.5%, indicating that only 14.5% of the words are correctly segmented into all their morphemes.

In comparison with the reference vocabularies, the accuracy of BETO's vocabulary is identical to that of *Wordpiece*_31 (14.54%) and nearly twice that of *BPE*_31 (8.7%). Regarding the average number of tokens used in each segmentation, it is noted that BETO and *Wordpiece*_31 tokenizers use the same number of tokens, while *BPE* uses nearly twice as many. These results suggest a similarity between BETO and *Wordpiece*_31 and a notable difference from *BPE*. Furthermore, since no improvement in accuracy is observed, it can be concluded that increasing the size of the training corpus for the tokenizers does not enhance the quality of the vocabularies they generate.

5.2. ERROR ANALYSIS

Error analysis was done using the error typology established in our previous work (García-Sierra et al., 2024), summarized as follows:

- Type 1: Under-segmentation errors occur when a very frequent word that has more than one morpheme is treated as a single token by the tokenizer.

- Type 2: Over-segmentation errors occur in infrequent words that contain frequent sub-words.

- Type 3: Errors of absence of the correct morpheme in the vocabulary, where the tokenizer uses more frequent sub-words that are not actual morphemes.

- Type 4: Errors in morpheme selection occur when the morpheme is in the vocabulary but is not selected by the tokenizer.

Table 7 presents the number of errors of each type found in BETO's vocabulary, and Table 8 provides examples of these errors. As shown in Table 7, the error typology of BETO's tokenizer closely resembles that of a *Wordpiece*-type tokenizer and differs from the one of a *BPE*, with a predominance of Type 1 under-segmentation errors and, to a lesser extent, Type 3 errors related to the absence of the morpheme in the vocabulary. Additionally, the total number of errors is the same for both BETO and *Wordpiece*_31.

| Tokenizer\Error | Type 1 | Type 2 | Type 3 | Type 4 | Total |
|---|---|---|---|---|---|
| *Wordpiece*_31 | 436 | 16 | 352 | 248 | 1.052 |
| BETO | 462 | 16 | 346 | 228 | 1.052 |
| *BPE*_31 | 98 | 85 | 591 | 350 | 1.124 |

**Table 7**. Quantitative results of BETO error analysis compared to *Wordpiece* and *BPE* with the same vocabulary size.

| Type 1 | Type 2 | Type 3 | Type 4 |
|---|---|---|---|
| ambos | néctar | deshago | inventada |
| both | nectar | (I) undo | invented |
| [amb, o, s] ✓ | [néctar] ✓ | [des, hag, o] ✓ | [invent, a, da] ✓ |
| [ambos] **X** | [n, éc, tar] **X** | [desha, go]* **X** | [inventa, da] **X** |
| contar | urbe | duerman | decoraciones |
| to count | city | (they) sleep | decorations |
| [cont, ar] ✓ | [urbe] ✓ | [duerm, a, n] ✓ | [decor, a, cion, es] ✓ |
| [contar] **X** | [ur, be] **X** | [duerma, n]* **X** | [decora, ciones] **X** |

**Table 8**. Examples of errors of each type in BETO. In the first line of each cell, the word to be segmented is indicated. In the second, the English translation can be found. In the third, the correct segmentation is shown. In the last one, BETO's tokenization is shown. In the word "deshago" the correct token "##hag" is not in BETO's vocabulary. In the word "duerman" the correct token "duerm" is also not in the tokenizers's vocabulary

## 6. DISCUSSION ON BETO'S TOKENIZATION ALGORITHM

Evaluation results suggest that BETO employs a *Wordpiece*-type segmentation algorithm. The only discrepancy with this conclusion appears in the morphological relevance results for prefixes and suffixes, where BETO and *BPE* yield nearly identical values. To verify these findings, we re-examined the evaluation process and confirmed that it had been done properly.

One possible explanation for the observed similarity between *BPE* and BETO in the relevance criterion could be that the tokenizer was initially trained using a *BPE* algorithm, but was switched to *Wordpiece* during the inference phase. This would also account for the inconsistency between the authors' claim of using *BPE* and the tokenizer configuration indicating the use of *Wordpiece*. However, we have deemed this explanation unlikely. Changing the tokenizer between training and inference phases is complex and would require more than just altering the algorithm's name in the model configuration file. It would also need modifications in BETO's repository structure, excluding files such as 'merges.txt' and adjustments to the token indicators. Therefore, if such a change in the segmentation algorithm were intentional, it would need to be explicitly indicated and justified.

Having ruled out the possibility of dual usage of *BPE* and *Wordpiece*, we conducted an additional experiment comparing the similarity of the prefix sets in vocabularies generated by *BPE* and BETO, and between *Wordpiece* and BETO. The same metrics of precision, coverage, and F1-score used for evaluating the morphological relevance criterion were applied. The results, presented in Table 9, indicate that the similarity between BETO and *Wordpiece* in terms of prefixes is greater than the similarity between BETO and *BPE*, with a significant difference observed in the F1 score. This finding supports the hypothesis that the tokenizer used in BETO is consistently a *Wordpiece* algorithm.

|  | **Precisión (%)** | **Cobertura (%)** | **F1 (%)** |
|---|---|---|---|
| BETO – *BPE*_31 | 97.78 | 80.0 | 88.0 |
| BETO – *Wordpiece*_31 | 98.15 | 96.36 | 97.25 |

**Table 9**. Similitud entre los prefijos de los vocabularios de BETO y *BPE*_31 y entre los de BETO y *Wordpiece*_31

## 7. SUMMARY, CONCLUSSIONS AND FUTURE WORK

In this work, we have addressed the problem of evaluating the quality of vocabularies in current large language models by examining the vocabulary of BETO's, an Spanish language model. These vocabularies are typically generated automatically by tokenizers such as *BPE*, *Wordpiece*, or *Unigram*, which are based on statistical strategies. The vocabulary is a key component for the effectiveness and reliability of a language model because it contains the linguistic units that the model is capable of recognizing and combining. Although statistical tokenizers are highly effective and do not require labeled corpora to build multilingual vocabularies for large language models, they generate subword vocabularies that do not always correspond to linguistic units such as morphemes or words. Measuring the quality of a vocabulary can help determine its impact on the performance of the language model. Furthermore, if the impact is significant, quality measures can facilitate improvements in the vocabularies.

In this work, we evaluated the vocabulary of BETO, a Spanish language model based on the BERT architecture. We employed an evaluation method developed in our previous work (García-Sierra et al., 2024). The objectives were to verify whether the findings from our previous work hold true regarding: (i) the low quality of vocabularies generated by segmentation algorithms used in large language models (specifically *BPE* and *Wordpiece*), (ii) the independence between the size of the training corpus for these segmentation algorithms and the quality of their vocabularies, (iii) the types of errors present in the vocabularies, and (iv) whether it is possible to determine the segmentation algorithm used by BETO's tokenizer.

From the results obtained in the evaluation of BETO's vocabulary and tokenizer, we can conclude that:

1. As with the tokenizers trained and evaluated in our previous work, the morphological quality of BETO's vocabulary is very low because it includes limited morphological knowledge of Spanish. Only in the case of prefix coverage BETO does achieve 90%, and for suffixes, it reaches 73%. For the remaining precision and coverage values for prefixes, suffixes, and stems, the results are low or very low.

2. Regarding the types of errors from statistical tokenizers—*BPE*, *Wordpiece*, or *Unigram*—we observe that BETO exhibits all these errors, with a predominance of Type 1 errors (under-segmentation) and Type 3 errors (absence of morphemes in the vocabulary), which are characteristic of a *Wordpiece* tokenizer.

3. Concerning the influence of corpus size on the quality of vocabularies generated by statistical segmentation algorithms, we found in BETO that using a corpus 500 times larger (300,000,000 sentences vs. 600,000 sentences) does not improve the quality of the generated vocabulary.

4. Finally, using the results from the evaluation of BETO's vocabulary quality, it can be inferred that the tokenizer used for its creation was *Wordpiece*, which matches the tokenizer indicated in the public version of the model and differs from the one claimed by BETO's authors.

Having a model and method for evaluating the quality of vocabularies in large language models opens the door to studying the question: to what extent does the morphological quality of vocabularies used by large language models influence their effectiveness and reliability? Our current and future research is focused on addressing this question.

Author Contribution Statement

Óscar García-Sierra: Research, Writing - Original Draft.

Ana Fernández-Pampillón Cesteros: Supervision, Methodology, Writing - Review and Editing.

Miguel Ortega-Martín: Supervision, Writing - Review and Editing.

Acknowledgments

We would like to thank our colleagues at Dezzai.

This work was supported by the Ministry of Science and Innovation project PID2022-140897OB-I00 ROBOT-TALK.

REFERENCES

Bostrom, K., y Durrett, G. (2020). Byte pair encoding is suboptimal for language model pretraining. *arXiv Preprint arXiv:2004. 03720*.

Cañete, J., Chaperon, G., Fuentes, R., Ho, J.-H., Kang, H., y Pérez, J. (2023). Spanish pre-trained bert model and evaluation data. *arXiv Preprint arXiv:2308. 02976*.

Church, K. W. (2020). Emerging trends: Subwords, seriously? *Natural Language Engineering*, *26*(3), 375–382.

Devlin, J., Chang, M.-W., Lee, K., y Toutanova, K. (2019). BERT: Pre-training of Deep Bidirectional Transformers for Language Understanding. *arXiv [Cs.CL]*. Retrieved from http://arxiv.org/abs/1810.04805

Fang, H., Ostendorf, M., Baumann, P., y Pierrehumbert, J. (2015). Exponential language modeling using morphological features and multi-task learning. *IEEE/ACM Transactions on Audio, Speech, and Language Processing*, *23*(12), 2410–2421.

Friedman, R. (2023). Tokenization in the Theory of Knowledge. *Encyclopedia, 3*(1), 380-386.

Hofmann, V., Pierrehumbert, J., y Schütze, H. (2021). Superbizarre is not superb: Derivational morphology improves BERT's interpretation of complex words. *Proceedings of the 59th Annual Meeting of the Association for Computational Linguistics and the 11th International Joint Conference on Natural Language Processing (Volume 1: Long Papers)*, 3594–3608.

Jabbar, H. (2024) MorphPiece: A Linguistic Tokenizer for Large Language Models. Retrieved from https://arxiv.org/html/2307.07262v2

Kudo, T., y Richardson, J. (2018). Sentencepiece: A simple and language independent subword tokenizer and detokenizer for neural text processing. *arXiv Preprint arXiv:1808. 06226*.

Lan, Z., Chen, M., Goodman, S., Gimpel, K., Sharma, P., y Soricut, R. (2019). Albert: A lite BERT for self-supervised learning of language representations. *arXiv preprint arXiv:1909.11942*.

Liu, Y., Ott, M., Goyal, N., Du, J., Joshi, M., Chen, D., … Stoyanov, V. (2019). RoBERTa: A Robustly Optimized BERT Pretraining Approach. *arXiv [Cs.CL]*. Retrieved from http://arxiv.org/abs/1907.11692

García-Sierra, Ó., Fernández-Pampillón, A, & Ortega-Martín, M. (2024). Evaluación morfológica de los vocabularios de subpalabras utilizados por los grandes modelos de lenguaje. Revista Española de Lingüística, 54(1), 103-130.

Park, K., Lee, J., Jang, S., y Jung, D. (2020). An Empirical Study of Tokenization Strategies for Various Korean NLP Tasks. *arXiv Preprint arXiv:2010. 02534*.

Radford, A., Narasimhan, K., Salimans, T., y Sutskever, I. (2018). *Improving language understanding by generative pre-training*. https://paperswithcode.com/paper/improving-language-understanding-by

Sennrich, R., Haddow, B., y Birch, A. (2015). Neural machine translation of rare words with subword units. *arXiv Preprint arXiv:1508. 07909*.

Schuster, M., y Nakajima, K. (2012). Japanese and korean voice search. *2012 IEEE International Conference on Acoustics, Speech and Signal Processing (ICASSP)*, 5149–5152. IEEE.

Song, X., Salcianu, A., Song, Y., Dopson, D., y Zhou, D. (2020). Fast *Wordpiece* tokenization. *arXiv preprint arXiv:2012.15524*.

Suárez, P. J. O., Sagot, B., y Romary, L. (2019). Asynchronous pipeline for processing huge corpora on medium to low resource infrastructures. *7th Workshop on the Challenges in the Management of Large Corpora (CMLC-7)*. Leibniz-Institut für Deutsche Sprache.

Suárez, P. J. O., Romary, L., y Sagot, B. (2020). A monolingual approach to contextualized word embeddings for mid-resource languages. *arXiv Preprint arXiv:2006. 06202*.

Van der Wouden, T. (1990). Celex: Building a multifunctional polytheoretical lexical data base. *Proceedings of BudaLex*, *88*, 363–373.

Vaswani, A., Shazeer, N., Parmar, N., Uszkoreit, J., Jones, L., Gomez, A. N., … Polosukhin, I. (2017). Attention is all you need. *Advances in Neural Information Processing Systems*, *30*.

Wu, Y., Schuster, M., Chen, Z., Le, Q. V., Norouzi, M., Macherey, W. y Dean, J. (2016). Google's neural machine translation system: Bridging the gap between human and machine translation. *arXiv Preprint arXiv:1609. 08144*.

# EVALUACIÓN MORFOLÓGICA DEL VOCABULARIO DE SUBPALABRAS DEL MODELO DEL LENGUAJE BETO

ÓSCAR GARCÍA-SIERRA[11]
*Universidad Complutense de Madrid; dezzai*

ANA FERNÁNDEZ-PAMPILLÓN CESTEROS[12]
*Universidad Complutense de Madrid*

MIGUEL ORTEGA-MARTÍN[13]
*Universidad Complutense de Madrid; dezzai*

*Resumen*

Los grandes modelos neuronales del lenguaje (LLM) utilizan, para la segmentación en palabras y morfemas de los textos que procesan, algoritmos de segmentación estadística en vez de algoritmos basados en reglas lingüísticas. La razón es que es que son mucho más eficaces y, sin intervención humana, son capaces de construir el necesario vocabulario de palabras y subpalabras. Sin embargo, las subpalabras que obtienen no siempre se corresponden con morfemas, lo que podría afectar negativamente al rendimiento de los modelos, aunque aún falta por determinar en qué medida. En trabajos anteriores, propusimos un método para evaluar la calidad morfológica de los vocabularios. En este artículo presentamos la aplicación del método de evaluación al vocabulario generado por el segmentador del modelo de lenguaje BETO. De dicha evaluación concluimos que la calidad del vocabulario de BETO es muy baja y el tipo de errores que presenta es confirman los resultados de nuestro trabajo anterior, que entrenar el segmentador con un corpus más extenso de 300 millones de frases no parece que mejore la calidad morfológica del vocabulario, y, finalmente, que el vocabulario de BETO parece que se ha generado con un segmentador tipo *Wordpiece*, como indica su configuración, y no *BPE*, como señalan los autores.

[11] oscarg02@ucm.es; https://orcid.org/0000-0002-8828-7338
[12] apampi@filol.ucm.es; https://orcid.org/0000-0002-6606-0159
[13] m.ortega@ucm.es; https://orcid.org/0000-0002-1880-5048

*Palabras clave*: segmentación; morfemas; subpalabras; grandes modelos del lenguaje; lengua española.

1. INTRODUCCIÓN

La segmentación (*tokenization* en inglés) es la división de un texto en unidades más pequeñas llamadas tokens (Friedman, 2023), que pueden ser morfemas, palabras, signos de puntuación, caracteres, etc. Entre otras utilidades, la segmentación es imprescindible para crear, a partir de conjuntos de textos, los conjuntos de datos para entrenar las redes neuronales artificiales, ya que estos datos representan las unidades discretas de información a partir de las cuales se construyen los modelos del lenguaje (Friedman, 2023).

En el entrenamiento de las redes neuronales actuales del tipo *Transformer* (Vaswani et al., 2017) para generar un modelo de lenguaje se utilizan vocabularios donde los tokens son palabras y subpalabras. Una subpalabra es una parte de una palabra que aparece con frecuencia significativa en el texto segmentado y que no necesariamente corresponde a un morfema. Por ejemplo, el modelo *BERT* (Devlin et al., 2019) y su versión en español, BETO (Cañete et al., 2023), utilizan vocabularios de treinta y un mil tokens*,* aproximadamente.

Entre los algoritmos de segmentación utilizados para generar los vocabularios de los modelos del lenguaje destacan tres por ser los que se utilizan mayoritariamente: *Wordpiece* (Schuster, 2012; Wu, 2016), utilizado, por ejemplo, por BERT; *Byte-Pair Encoding (BPE)* (Sennrich, 2015), empleado por la familia *GPT* (Radford et al., 2018) o por *RoBERTa* (Liu et al., 2019); y *Unigram*, utilizado por *Albert* (Lan et al., 2019).

Los algoritmos que se emplean actualmente para generar los mencionados vocabularios se fundamentan en la frecuencia con la que ciertas cadenas de caracteres aparecen en el conjunto de datos de entrenamiento. Es decir, se basan en criterios puramente estadísticos, lo que produce, como resultado, que las palabras con una frecuencia relevante en el corpus no se segmenten y sean un solo token del vocabulario, y, sin embargo, las palabras menos frecuentes se dividan en varias subpalabras con mayor frecuencia de aparición. Por ejemplo, el segmentador del modelo de lenguaje *BETO* no segmenta la palabra *población* -aun cuando está formada por la raíz *pobla* y el morfema

*##ción-* y lo considera como un solo token; sin embargo, segmenta la palabra *perduración* en los tokens *perd*, y *##uración* y la palabra *oscilación* en *oscila,*y *##cion*.

Frente a los segmentadores simbólicos basados en reglas lingüísticas, los segmentadores estadísticos presentan la ventaja de que son capaces de generar vocabularios compuestos por decenas de miles de tokens a partir de extensos corpus de texto sin necesidad de supervisión humana, lo cual resulta especialmente útil en el desarrollo de modelos multilingües. Sin embargo, como demuestran diversos estudios (Church, 2020; Bostrom y Durret, 2020; Hoffman et al., 2021; Park, 2020), tienen el inconveniente de que la segmentación en subpalabras no se corresponde en gran medida con los morfemas de la lengua, lo que resulta en una pérdida del conocimiento lingüístico sobre el cual se construyen los modelos del lenguaje y que podría estar disminuyendo la eficacia y fiabilidad de los modelos. Por ejemplo, en inglés, Hoffman et al. (2021) comprueban que un modelo BERT con un vocabulario de morfemas mejora los resultados de otro modelo BERT con un vocabulario generado por un segmentador *BPE* en las tareas de análisis de sentimiento y predicción de tópicos. Jabbar (2024), también en inglés, compara un modelo GPT con *BPE* con otro GPT con un segmentador propio llamado MorphPiece, y el segundo supera al primero en una ampla variedad de tareas.

En nuestro trabajo anterior hemos propuesto una solución a la cuestión de cómo evaluar la calidad morfológica, basándonos en los resultados de los trabajos previos (García-Sierra et al., 2024). Así, definimos la *calidad morfológica del vocabulario de un modelo del lenguaje en una lengua* como su grado de semejanza respecto de un vocabulario real de morfemas de esa misma lengua y establecimos tres criterios para medirlo: la relevancia, la coherencia y la corrección morfológica. Cada uno de estos criterios mide un aspecto del grado de semejanza entre los vocabularios de forma que, (i), la relevancia morfológica se refiere a la proporción de morfemas reales de la lengua que contiene el vocabulario del LLM, (ii), la coherencia morfológica mide la frecuencia con la que las palabras que comparten algún morfema comparten también un token que sea dicho morfema y, (iii), la corrección morfológica mide la capacidad del segmentador del LLM para utilizar los morfemas reales de la lengua (y no otras subpalabras del vocabulario) para segmentar correctamente las palabras de los conjuntos de datos de evaluación. Además, propusimos un método para medir cada criterio y lo aplicamos a la evaluación de los tres algoritmos de subpalabras *BPE*, *Wordpiece* y *Unigram* trabajando con textos del español que es una lengua con una complejidad morfológica mayor que la de la lengua inglesa. Los resultados indicaron que, para el español, los tres algoritmos tienen una calidad morfológica muy baja y que, aunque se aumente el tamaño de los vocabularios no mejora su calidad morfológica. También indicaron la existencia de algunos tipos de errores intrínsecos de los segmentadores. Así, *Unigram* y *Wordpiece*, a diferencia de *BPE* tienden a no segmentar las palabras; *BPE*, por su parte, genera vocabularios con más ausencia de morfemas que *Unigram* y *Wordpiece*, por lo que las segmentaciones son peores especialmente respecto a la relevancia y la corrección morfológica.

En este artículo presentamos los resultados de la aplicación del modelo y método de evaluación al caso real del vocabulario generado por el segmentador del modelo de lenguaje BETO disponible públicamente[14]. BETO es un modelo de lenguaje para el español de tipo BERT creado en 2019 y que está basado en el codificador de la arquitectura *Transformer*, entrenado solamente con textos en español. En concreto el corpus de entrenamiento está formado por trescientos millones de frases, frente a las seiscientas mil que tenía el corpus utilizado para entrenar los segmentadores evaluados en nuestro trabajo anterior.

La evaluación de la calidad del vocabulario de BETO tiene interés porque, (i), es uno de los modelos de lenguaje más descargados en español de la plataforma de *Hugging Face*, (ii) permitirá determinar si el vocabulario que utiliza podría ser una posible limitación en el rendimiento del modelo y, de ser así, estudiar si la mejora del vocabulario conduce a una mejora significativa del modelo, y, (iii) permitirá verificar si se confirman los resultados obtenidos en los experimentos realizados en García-Sierra et al (2024) sobre la calidad de los vocabularios que generan los algoritmos de segmentación *Wordpiece* y *BPE*, y, más concretamente, verificar si la calidad de los vocabularios que se generan es baja, si se confirma la tipología de errores encontrada en cada tipo de segmentador, y, si entrenar BETO con un corpus más extenso repercute en una mayor calidad morfológica del vocabulario del segmentador.

El artículo se ha organizado de la forma siguiente: en la sección segunda definimos los objetivos del trabajo; en la sección tercera indicamos los posibles segmentadores que se han utilizado para generar el vocabulario de BETO y que se debe utilizar para aplicar BETO a las tareas de PLN; en la sección cuarta resumimos el procedimiento de evaluación de la calidad morfológica; en la sección quinta se describe la aplicación de dicho procedimiento a la evaluación de la calidad del vocabulario generado por el segmentador de BETO; en la sección sexta se presentan y discuten los resultados, primero los valores cuantitativos obtenidos para cada criterio de evaluación y, después, mediante un análisis de errores. En la séptima sección discutimos el algoritmo utilizado por el segmentador de BETO. En última sección, octava, se presenta un resumen, las conclusiones y el trabajo futuro.

---

[14] https://github.com/dccuchile/beto

## 2. OBJETIVOS DEL TRABAJO

El objetivo principal del trabajo es evaluar la calidad del vocabulario del modelo del lenguaje BETO para comprobar:

(i) si se verifican los resultados encontrados en nuestro trabajo anterior sobre la baja calidad de los vocabularios que generan los algoritmos de segmentación *BPE* y *Wordpiece*,

(ii) la independencia, a partir de un tamaño umbral, entre el tamaño del corpus de entrenamiento de estos segmentadores y la calidad de sus vocabularios.

(iii) el tipo de errores que presentan los vocabularios generados por estos algoritmos de segmentación, y,

(iv) si es posible determinar, a partir de la evaluación de la calidad del vocabulario de BETO, cuál es el algoritmo de segmentación que ha sido utilizado para su generación.

## 3. EL SEGMENTADOR DEL MODELO BETO

El segmentador del modelo de lenguaje BETO ha sido entrenado en un corpus de trescientos millones de frases, es decir, un corpus quinientas veces más extenso que el usado para entrenar los segmentadores de García-Sierra et al. (2024).

Sin embargo, con la información disponible actualmente no está claro qué tipo de segmentador que utiliza BETO. Según sus autores, el segmentador está basado en el algoritmo *BPE* (Cañete et al., 2023:3), pero esta información no se corresponde con la especificada en el sitio de descargas de la plataforma *Hugging Face*[15], en el archivo de configuración del segmentador[16] en el que se indica que el segmentador utilizado es de tipo *Wordpiece* (figura 1).

---

[15] https://huggingface.co/dccuchile/bert-base-spanish-wwm-uncased
[16] https://huggingface.co/dccuchile/bert-base-spanish-wwm-uncased/raw/main/tokenizer.json

```
"decoder": {
  "type": "WordPiece",
  "prefix": "##",
  "cleanup": true
},
"model": {
  "type": "WordPiece",
  "unk_token": "[UNK]",
  "continuing_subword_prefix": "##",
  "max_input_chars_per_word": 100,
  "vocab": {
```

**Figura 1.** Configuración del segmentador de BETO en la plataforma Hugging Face.

Nuestra intuición en este punto es que probablemente sea *Wordpiece* el tipo de segmentador de BETO en la versión actual del modelo no sólo por las indicaciones del archivo de configuración sino también porque, de tratarse de *BPE*, este debería incluir un archivo "merges.txt" que guardase las uniones de caracteres aprendidas durante el entrenamiento por orden de frecuencia, y este archivo no aparece en el repositorio de BETO en *Hugging Face*[17]. Sin embargo, teniendo en cuanta las indicaciones de los autores podría ocurrir que el vocabulario se hubiese, efectivamente, creado con el vocabulario generado por *BPE* y, en las indicaciones de configuración se estuviese indicando el uso de un tipo de segmentador diferente para la fase de inferencia lo cual tendría repercusión en el rendimiento del modelo puesto que se estarían usando subpalabras diferentes.

Nuestra hipótesis, en este sentido es que nuestro objetivo de evaluar el vocabulario de BETO ayudará a resolver esta incongruencia y a precisar el algoritmo utilizado por el segmentador mediante la comparación de los resultados que se obtengan de las métricas de calidad y la tipología de errores con los resultados obtenidos en nuestro trabajo anterior.

## 4. MÉTODO DE EVALUACIÓN DE LA CALIDAD MORFOLÓGICA DE LOS VOCABULARIOS DE LOS MODELOS DEL LENGUAJE

Nuestra propuesta es que el método de evaluación de la calidad morfológica se base en la medición del grado de similitud entre el vocabulario del modelo del lenguaje generado por un segmentador y el vocabulario real de morfemas de la lengua del modelo (García-Sierra et al., 2024). Proponemos que el grado de similitud depende de los tres aspectos o criterios de calidad -mencionados en la sección 1 de introducción-, la relevancia, coherencia y corrección morfológicas, teniendo en cuenta que: (i), la relevancia morfológica se refiere a la proporción de morfemas reales de la lengua del modelo del lenguaje que aprendidos por el segmentador, (ii), la coherencia morfológica mide la frecuencia con la que las palabras que comparten un morfema comparten también un token que se corresponde con dicho morfema y, (iii), la corrección morfológica mide la capacidad del segmentador para utilizar los morfemas reales de la lengua que ha aprendido para segmentar correctamente las palabras.

---

[17] https://huggingface.co/dccuchile/bert-base-spanish-wwm-uncased/tree/main

Para medir los tres criterios es necesario: (1) construir un conjunto de datos de evaluación para cada criterio, y, con ellos, (2), calcular el nivel de cumplimiento de cada criterio.

(1) Para la creación de los tres conjuntos de datos de evaluación tendremos en cuenta que:

i. el conjunto de datos para evaluar la relevancia morfológica de los vocabularios debe ser la lista de morfemas de la lengua, a ser posible divididos por su tipología: prefijo, sufijo y raíz. La tabla 1 describe nuestro conjunto de datos para el español.

| **Tipo de morfema** | **Total** | **Ejemplo** |
|---|---|---|
| Prefijos | 61 | ['des', 're', ...] |
| Sufijos | 175 | ['mos', 'ción', ...] |
| Raíces | 5,000 | ['dej', 'alt', ...] |

**Tabla 1.** Conjunto de datos de evaluación de la relevancia morfológica para el español (fuente: García Sierra et al., 2024)

ii. El conjunto de datos para el criterio de la coherencia morfológica debe ser una lista de pares palabra y morfema para todos los morfemas de la lengua. Esta lista se puede crear a partir del conjunto de datos anterior de morfemas y una lista, lo más completa posible, de palabras de la lengua. La tabla 2 describe nuestro conjunto de datos para el español.

| Tipo de morfema | Total de palabras | Ejemplo |
|---|---|---|
| Prefijos | 205.792 | [(*remodelación*, *re-*), …] |
| Raíces | 234.564 | [(*ubicásemos*, *ubic*), …] |
| Sufijos | 1.575.207 | [(*fijación*, *-ción*), …] |
| Clíticos | 3.920 | [(*dilo*, *lo*), …] |

**Tabla 2.** Conjunto de datos de evaluación de la coherencia morfológica para el español (fuente: García Sierra et al., 2024)

iii. El conjunto de datos para evaluar la corrección morfológica debe ser una lista de palabras etiquetadas con su(s) correspondiente(s) segmentación(es) morfológica(s). Para su construcción se puede tomar una muestra aleatoria de palabras de forma que, para asegurar la máxima representatividad, estén presentes todos los morfemas de la lengua al menos una vez y mantener la proporción de categorías gramaticales del diccionario de la lengua en cuestión. La tabla 3 describe el conjunto de datos construido por nosotros para el español.

| Categoría gramatical | Total | Ejemplos |
|---|---|---|
| Nombres | 481 | ['camiones': {'NOUN': ['camion', 'es']}] |
| Verbos | 445 | ['dejan': {'VERB': ['dej', 'an']}] |
| Adjetivos | 299 | ['adecuadas': {'ADJ': ['adecu', 'ad', 'a', 's']}] |
| Pronombres | 61 | ['estos': {'PRON': ['est', 'o', 's']}] |
| Adverbios | 59 | ['aquí': {'ADV': ['aquí']}] |
| Determinantes | 30 | ['mis': {'DET': ['mi', 's']}] |
| Preposiciones | 22 | ['desde': {'ADP': ['desde']}] |
| Conjunciones | 14 | ['pero': {'CONJ': ['pero']}] |

**Tabla 3**. Conjunto de datos de evaluación de la corrección morfológica para el español (fuente: García Sierra et al., 2024)

(2) Para calcular el nivel de cumplimiento de cada criterio se utilizarán las métricas siguientes:

i. Para la relevancia morfológica, las métricas tradicionales de *precisión, cobertura* y *valor F1* que miden la intersección entre los token del vocabulario del modelo del lenguaje y los morfemas de la lengua recogidos en el conjunto de datos de relevancia morfológica (ej. tabla 1).

ii. Para la coherencia morfológica, se calculan tres medidas para cada tipo de morfema (prefijo, sufijo, raíces, clíticos):

Medida 1 ("segmentación en un solo token"): es porcentaje, sobre el total de palabras del conjunto de datos de evaluación de la coherencia, del total de palabras que contienen el tipo de morfema y que, sin embargo, se segmentan como un solo *token* (es decir, que no se ha segmentado); esta medida permite reconocer un tipo de errores que denominamos de "infrasegmentación"[18].

Medida 2 ("el morfema no se reconoce"): porcentaje, sobre el total de palabras del conjunto de datos de evaluación de la coherencia, del total palabras que contienen el tipo de morfema y se segmentan como varios tokens, pero el morfema no se corresponde con ninguno de estos token de la segmentación; esta medida permite reconocer errores del tipo "sobresegmentación" o de "no uso del morfema".

Medida 3 ("el morfema se reconoce"): porcentaje, sobre el total de palabras del conjunto de datos de evaluación de la coherencia, del total palabras que contienen el morfema y se segmentan como varios tokens en las que morfema se corresponde con uno token de la segmentación. Esta última medida indica la capacidad del segmentador del modelo para "reconocer" ese tipo de morfemas y, por lo tanto, lo efectivo que es el vocabulario del modelo para reconocer ese tipo de morfemas (utilizando el segmentador).

iii. Para la corrección morfológica, se calcula en qué proporción de palabras respecto al total todos tokens de la segmentación de esa palabra del conjunto de datos de evaluación de la corrección, coinciden con los morfemas etiquetados. El rango de la corrección varía entre 0 y 1 (o 100% si se usan valores porcentuales), indicando con 0 que ninguna palabra se ha segmentado correctamente y 1 (o 100) indicando que todas las palabras han sido segmentadas correctamente.

---

[18] Sobre la tipología de errores de segmentación se puede consultar la página 124 de García-Sierra et al. (2024)

# 5. EVALUACIÓN DE LA CALIDAD DEL VOCABULARIO DEL MODELO BETO

Para evaluar la calidad del vocabulario de la versión pública del modelo del lenguaje BETO en *Hugging Face* utilizamos el modelo y método de evaluación descrito en el apartado anterior. Como hemos indicado en la sección tercera, el tipo de segmentador que utiliza BETO, según el archivo de configuración, es el segmentador *Wordpiece*, aunque los autores de BETO especifican que es *BPE*.

Los tres conjuntos de datos de evaluación que hemos utilizado han sido los construidos en nuestro trabajo previo publicado en García-Sierra et al., (2024). Los detalles del procedimiento de construcción pueden consultarse en dicha publicación. Los conjuntos de datos están disponibles públicamente y los algoritmos para el cálculo de las métricas también están disponibles públicamente[19]..

Además, los resultados de la calidad del vocabulario actual de BETO que hemos obtenido se presentan comparativamente respecto a los resultados de calidad de dos de los vocabularios generados en nuestro trabajo anterior con los algoritmos de segmentación *Wordpiece* y *BPE*. Estos vocabularios son de tamaño 31.000 tokens y fueron entrenados con el corpus Oscar de 600.000 frases disponible en *Hugging Face*[20]. Utilizaremos la nomenclatura *Wordpiece*_31 y *BPE*_31 para referirnos a estos dos segmentadores y consideraremos los dos vocabularios generados por *Wordpiece*_31 y *BPE*_31 como referencia o "línea base" respecto a la que comparamos los resultados de la evaluación del vocabulario de BETO. De esta forma podremos comprobar, entre otras cuestiones, si el uso de conjuntos de datos de mayor tamaño para el entrenamiento de los segmentadores mejora la calidad morfológica de los vocabularios, o si alguno de los vocabularios de referencia guarda similitud, respecto a su calidad con el vocabulario de BETO.

*5.1. Resultados y discusión*

5.1.1. *Relevancia morfológica*

La tabla 4 muestra los resultados de la calidad del vocabulario del modelo BETO y los vocabularios de referencia respecto el criterio de relevancia morfológica. Al igual que los vocabularios de referencia, el vocabulario de BETO obtiene una precisión muy baja, señal de que el funcionamiento estadístico de su algoritmo de segmentación no es capaz de aprender adecuadamente las unidades morfológicas del español. Los mejores valores se obtienen respecto a la cobertura; así, el vocabulario de BETO obtiene un 90% para los prefijos, casi un 73% para los sufijos y un 23% para las raíces. Esto indica que incluye en su vocabulario una proporción muy alta de prefijos, alta de sufijos y muy baja de raíces de la lengua española. Serán los otros dos criterios de coherencia y corrección los que indicarán si estos prefijos y sufijos aprendidos son utilizados realmente por el segmentador cuando se usa BETO en las inferencias. El valor F1 es muy bajo al ser la media armónica de precisión y cobertura y ser la precisión muy baja.

---

[19] https://github.com/ogarciasierra/spanish-subwords-evaluation
[20] https://huggingface.co/datasets/nthngdy/oscar-small

| Tipo | Totales | Segmentador | Precisión (%) | Cobertura (%) | F1 (%) |
|---|---|---|---|---|---|
| Prefijos | 61 | BETO | 0,18 | 90,16 | 0,37 |
| | | *Wordpiece*_31 | 0,17 | 88,52 | 0,35 |
| | | *BPE*_31 | 0,18 | 90,16 | 0,37 |
| Sufijos | 175 | BETO | 0,41 | 72,99 | 0,81 |
| | | *Wordpiece*_31 | 0,41 | 73,56 | 0,82 |
| | | *BPE*_31 | 0,41 | 72,41 | 0,81 |
| Raíces | 5,000 | BETO | 3,84 | 23,78 | 6,61 |
| | | *Wordpiece*_31 | 8,59 | 9,97 | 9,23 |
| | | *BPE*_31 | 7,75 | 8,99 | 8,32 |

**Tabla 4.** Resultados de la evaluación de la relevancia morfológica del segmentador de BETO y comparación con los de *Wordpiece* y *BPE*.

Respecto a la comparativa del vocabulario de BETO con los vocabularios de referencia de *Wordpiece*_31 y *BPE*_31 encontramos que:

(1) respecto a los prefijos y sufijos, los valores de las tres métricas son prácticamente iguales entre el vocabulario de BETO y el vocabulario *BPE*_31 y, en media, muy semejantes entre los tres. Esto podría indicar que, al menos respecto a los prefijos y sufijos, el segmentador que se ha usado en BETO podría ser de tipo *BPE*. Sin embargo,

(2), existe una diferencia de más de diez puntos en los morfemas de tipo raíz entre los valores de cobertura del vocabulario de BETO (24%) y los vocabularios de referencia (10% y 9%). Este resultado podría indicar que, entrenar el segmentador de un modelo con un corpus más extenso, mejora la cobertura de las raíces, aunque no la de los prefijos y sufijos.

Respecto a la comparación del algoritmo de segmentación de BETO con los algoritmos *Wordpiece* y *BPE*, observamos que los resultados entre BETO y *BPE* son iguales. Sin embargo, los resultados entre *Wordpiece* y *BPE* son muy semejantes por lo que, valiéndonos solo de este criterio, no podemos precisar cuál de los dos algoritmos es el utilizado por el segmentador de BETO.

*5.1.2. Coherencia morfológica*

La tabla 5 muestra los resultados del uso del vocabulario que hace el segmentador de BETO y los vocabularios de referencia para la evaluación del criterio de coherencia morfológica.

| **Categoría** | **Vocabulario** | **Total tokens** | **Segmentación errónea en un solo token (%)** | **Varios token-correctos (%)** | **Varios token - incorrecto (%)** |
|---|---|---|---|---|---|
| Prefijos | BETO | 205.792 | 0.87 | 12.20 | 86.93 |
| | *Wordpiece*_31 | | 0,71 | 13,45 | 85,84 |
| | *BPE*_31 | | 0,04 | 65,86 | 34,10 |
| Raíces | BETO | 234.584 | 1.95 | 17.27 | 80.78 |
| | *Wordpiece*_31 | | 1,64 | 16,03 | 82,33 |
| | *BPE*_31 | | 0,40 | 4,66 | 94,94 |
| Sufijos | BETO | 1.575.207 | 1.01 | 13.96 | 85.03 |
| | *Wordpiece*_31 | | 0,74 | 15,17 | 84,09 |
| | *BPE*_31 | | 0,11 | 10,08 | 89,81 |
| Clíticos | BETO | 3.291 | 11.43 | 53.94 | 33.63 |
| | *Wordpiece*_31 | | 5,61 | 61,39 | 33,00 |
| | *BPE*_31 | | 0,69 | 48,58 | 50,73 |
| Totales | BETO | 2.019.504 | 1,12 | 14,25 | 84,63 |
| | *Wordpiece*_31 | | 0,86 | 15,18 | 83,96 |
| | *BPE*_31 | | 0,14 | 15,21 | 84,66 |

**Tabla 5.** Resultados de la evaluación de la coherencia morfológica

Como se observa, el segmentador de BETO tiene unos valores de coherencia bastante bajos: sólo en un 12% de las veces utiliza siempre los mismos prefijos para segmentar las palabras que los contienen, un 17% en las raíces, un 13% en los sufijos, y se incrementa en casi un 54% en los clíticos.

Respecto a la comparación con los vocabularios de referencia, excepto en el caso de los clíticos, los resultados de coherencia de BETO son, a diferencia de lo que observamos en la relevancia, similares a los de *Wordpiece*_31 y significativamente diferentes a *BPE*. Además, los resultados de coherencia del vocabulario de BETO son ligeramente peores que *Wordpiece*_31 excepto en las raíces por lo que no parece que entrenar el segmentador con un corpus más extenso implique una mejora de la coherencia morfológica.

5.1.3. *Corrección morfológica.*

La tabla 6 muestra los resultados de la evaluación del vocabulario de BETO respecto al criterio de corrección morfológica junto con los de los dos vocabularios de referencia.

| Segmentador | Corrección (%) | Token utilizados de media por palabra |
|---|---|---|
| BETO | 14,54 | 1,71 |
| *Wordpiece*_31 | 14,54 | 1,75 |
| *BPE*_31 | 8,69 | 2,39 |

**Tabla 6.** Resultados de la corrección morfológica, token utilizados de media por palabra y comparación con los resultados de *Wordpiece* y *BPE*.

En este caso los valores de la corrección en el vocabulario generado por el segmentador de BETO son muy bajos, un 14,5%, lo que significa que sólo en un 14,5% de las palabras son segmentadas correctamente en todos sus morfemas.

Respecto a los vocabularios de referencia, observamos que la corrección en el uso del vocabulario de BETO es idéntica a la que realiza *Wordpiece*_31 (14,54% de corrección) y casi el doble que la del vocabulario de *BPE*_31 (8,7%). Respecto a la media de token utilizados en cada segmentación por los diferentes segmentadores, se observa que es la misma en el segmentador de BETO y *Wordpiece*_31 mientras que *BPE* utiliza casi el doble de tokens. Estos resultados indican una similitud entre BETO y *Wordpiece*_31 y una diferencia notable con *BPE*. Además, al no observarse mejora en la corrección, se puede afirmar que también en este caso aumentar el tamaño del corpus de entrenamiento de los segmentadores no mejora la calidad de los vocabularios que generan.

## 5. 2. ANÁLISIS DE ERRORES

El análisis de errores se ha realizado utilizando la tipología de errores que establecimos de nuestro anterior trabajo (García-Sierra et al., 2024) y que, en resumen, es la siguiente:

(i) Tipo 1. Errores de infrasegmentación, ocurre cuando no se segmenta alguna palabra muy frecuente en el corpus de entrenamiento del segmentador que contiene morfemas que sí están en el vocabulario pero que son menos frecuentes.

(ii) Tipo 2. Errores de sobresegmentación ocurre en las palabras poco frecuentes que contienen subpalabras frecuentes.

(iii) Tipo 3. Errores de ausencia del morfema correcto en el vocabulario, por lo que el segmentador utiliza subpalabras más frecuentes que no son morfemas.

(iv) Tipo 4. Errores en la selección del morfema, que ocurren cuando el morfema sí está en el vocabulario pero el segmentador no lo selecciona.

La tabla 7 muestra el número de errores de cada tipo que se han encontrado en el vocabulario de BETO y la tabla 8 muestra algunos ejemplos de estos errores

Como se observa en la tabla 7, la tipología de errores del segmentador de BETO es muy similar a la de un segmentador de tipo *Wordpiece* y diferente de *BPE*, predominando los errores de tipo 1, de infrasegmentación, y, en menor medida, los de tipo 3, de ausencia del morfema en el vocabulario. Además, se observa que el total de errores es el mismo en BETO y en *Wordpiece*_31.

| Segmentador\Error | Tipo 1 | Tipo 2 | Tipo 3 | Tipo 4 | Total |
|---|---|---|---|---|---|
| *Wordpiece*_31 | 436 | 16 | 352 | 248 | 1.052 |
| BETO | 462 | 16 | 346 | 228 | 1.052 |
| *BPE*_31 | 98 | 85 | 591 | 350 | 1.124 |

**Tabla 7.** Resultados cuantitativos del análisis de errores de BETO en comparación con los de *Wordpiece* y *BPE* con su mismo tamaño de vocabulario.

| Tipo 1 | Tipo 2 | Tipo 3 | Tipo 4 |
|---|---|---|---|
| ambos [amb, o, s] ✓ [ambos] X | néctar [néctar] ✓ [n, éc, tar] X | deshago [des, hag, o] ✓ [desha, go]* X | inventada [invent, a, da] ✓ [inventa, da] X |
| contar: | urbe | duerman | decoraciones |

| [cont, ar] ✓ | [urbe] ✓ | [duerm, a, n] ✓ | [decor, a, cion, es] ✓ |
| [contar] X | [ur, be] X | [duerma, n]* X | [decora, ciones] X |

**Tabla 8**. Ejemplos de errores de cada de tipo en BETO. En la primera línea de cada celda, se indica la palabra a segmentar. En la segunda, se muestra la segmentación correcta. En la tercera, se muestra la segmentación de BETO. En la palabra "deshago" el token correcto "##hag" no está en el vocabulario de BETO. En la palabra "duerman" el token correcto "duerm" tampoco está en el vocabulario del segmentador.

## 6. DISCUSIÓN SOBRE EL ALGORITMO DE SEGMENTACIÓN DE BETO

Los resultados de la evaluación parecen indicar que el modelo BETO utiliza un algoritmo de segmentación de tipo *Wordpiece*. Únicamente los resultados de la relevancia morfológica para el caso de los prefijos y sufijos parecen contradecir esta conclusión puesto que *BPE* y BETO obtienen, exactamente o prácticamente, los mismos valores. Para verificar estos valores, se llevó a cabo una nueva revisión del proceso de evaluación y de los resultados, comprobando que efectivamente se habían realizado correctamente.

Una posible explicación a esta similitud discrepante entre *BPE* y BETO, sólo en el criterio de relevancia, es que el segmentador se hubiese entrenado con un algoritmo *BPE* y posteriormente, para la fase de inferencia, se hubiese cambiado por *Wordpiece*. Esto también explicaría la discrepancia entre los autores, que afirman haber usado *BPE* y la configuración del segmentador que indica el uso de *Wordpiece*. Hemos rechazado esta explicación como posible porque un cambio de segmentador en los procesos de entrenamiento e inferencia es complicado: no sólo es necesario cambiar el nombre del algoritmo de segmentación en el fichero de configuración del modelo, sino que también conllevaría pasos adicionales, como modificar la estructura del repositorio de BETO, archivos como "merges.txt" o los caracteres que se utilizan para señalar si un token es o no comienzo de palabra. Así, entendemos que, si fuese intencionado el cambio de algoritmo de segmentación en BETO en las fases de entrenamiento e inferencia, sería necesario que se indicase y justificase explícitamente.

Por último, desechada la posibilidad del doble uso de *BPE* y *Wordpiece*, se realizó un nuevo experimento para comparar la similitud entre el conjunto de prefijos de los vocabularios generados por *BPE* y BETO y la similitud entre el conjunto de prefijos de *Wordpiece* y BETO. Para ello se utilizaron las mismas métricas de precisión, cobertura y F1 utilizadas para evaluar el criterio de relevancia morfológica. Los resultados, que se incluyen en la tabla 9, muestran que, para el caso de los prefijos, la similitud es mayor entre BETO y *Wordpiece* que entre BETO y *BPE* siendo la diferencia significativa respecto el valor F1. Este nuevo resultado parece confirmar la hipótesis de que el segmentador utilizado en el modelo BETO es siempre un algoritmo *Wordpiece*.

|  | **Precisión (%)** | **Cobertura (%)** | **F1 (%)** |
|---|---|---|---|
| BETO – *BPE*_31 | 97.78 | 80.0 | 88.0 |
| BETO – *Wordpiece*_31 | 98.15 | 96.36 | 97.25 |

**Tabla 9**. Similitud entre los prefijos de los vocabularios de BETO y *BPE*_31 y entre los de BETO y *Wordpiece*_31.

## 7. RESUMEN, CONCLUSIONES Y TRABAJO FUTURO

En este trabajo hemos abordado el problema de evaluar la calidad de los vocabularios de los actuales grandes modelos del lenguaje estudiando el caso del vocabulario del modelo BETO en español. Estos vocabularios son, habitualmente, generados automáticamente por segmentadores tipo *BPE*, *Wordpiece* o *Unigram* basados en estrategias estadísticas. El vocabulario es un componente clave para la eficacia y fiabilidad de un modelo del lenguaje porque contiene las unidades lingüísticas que el modelo es capaz de reconocer y combinar. Pese a que los segmentadores estadísticos son altamente eficaces y no necesitan corpus etiquetados para construir los vocabularios multilingües de los grandes modelos del lenguaje, presentan el problema de que generan vocabularios de subpalabras que no siempre se corresponden con unidades lingüísticas de tipo morfema o palabra. El poder medir cuánto de bueno es un vocabulario puede ayudar a determinar el impacto de la calidad de éste en el funcionamiento del modelo del lenguaje. Asimismo, si el impacto es significativo, las medidas de calidad facilitarán la mejora de los vocabularios.

En este trabajo hemos evaluado el vocabulario del modelo BETO, un modelo de lenguaje en español basado en la arquitectura BERT. Para ello, hemos utilizado un método de evaluación que desarrollamos en nuestros anteriores trabajos (García Sierra et al, 2024). Los objetivos son comprobar si se verifican los resultados encontrados en nuestro trabajo anterior sobre: (i) la baja calidad de los vocabularios que generan los algoritmos de segmentación utilizados en los grandes modelos del lenguaje (específicamente *BPE* y *Wordpiece*), (ii) la independencia entre el tamaño del corpus de entrenamiento de estos algoritmos de segmentación y la calidad de sus vocabularios, (iii) la tipología de errores que presentan los vocabularios, y, (iv), si es posible determinar el algoritmo de segmentación del modelo BETO.

De los resultados obtenidos en la evaluación de la calidad del vocabulario y segmentador de BETO podemos concluir que:

(1) como ocurría con los segmentadores entrenados y evaluados en nuestro trabajo previo, la calidad morfológica del vocabulario de BETO es muy baja porque incluye poco conocimiento morfológico del español. Únicamente en el caso de la cobertura de los prefijos BETO logra un 90% y en los sufijos un 73%. En el resto de los valores de precisión y cobertura de prefijos, sufijos y raíces los valores son bajos o muy bajos.

(2) Respecto al tipo de errores de los segmentadores estadísticos -*BPE*, Wordpice o *Unigram*- comprobamos que en BETO se observan todos ellos y predominan el tipo de errores 1 de infrasegmentación y 3 de ausencia de morfemas en el vocabulario propios del segmentador *Wordpiece*.

(3) Respecto a la influencia del tamaño del corpus en la calidad de los vocabularios que generan los algoritmos de segmentación estadísticos, se ha comprobado en BETO que el uso de un corpus quinientas veces más extenso (300.000.000 frases vs. 600.000 frases) no mejora la calidad del vocabulario generado. Finalmente,

(4) utilizando los resultados de la evaluación de la calidad del vocabulario de BETO se puede inferir que el segmentador que se ha utilizado para su creación ha sido *Wordpiece*, que es el que aparece en la versión pública del modelo y diferente de la que indican los autores del modelo BETO.

Disponer de un modelo y método de evaluación de la calidad de los vocabularios de los grandes modelos del lenguaje abre la puerta a abordar el estudio de la cuestión ¿hasta qué punto la calidad morfológica de los vocabularios utilizados por los grandes modelos del lenguaje influye en su eficacia y fiabilidad? Nuestra investigación actual y futura se enfoca en abordar esta cuestión.

REFERENCIAS

Bostrom, K., y Durrett, G. (2020). Byte pair encoding is suboptimal for language model pretraining. *arXiv Preprint arXiv:2004. 03720*.

Cañete, J., Chaperon, G., Fuentes, R., Ho, J.-H., Kang, H., y Pérez, J. (2023). Spanish pre-trained bert model and evaluation data. *arXiv Preprint arXiv:2308. 02976*.

Church, K. W. (2020). Emerging trends: Subwords, seriously? *Natural Language Engineering*, *26*(3), 375–382.

Devlin, J., Chang, M.-W., Lee, K., y Toutanova, K. (2019). BERT: Pre-training of Deep Bidirectional Transformers for Language Understanding. *arXiv [Cs.CL]*. Retrieved from http://arxiv.org/abs/1810.04805

Fang, H., Ostendorf, M., Baumann, P., y Pierrehumbert, J. (2015). Exponential language modeling using morphological features and multi-task learning. *IEEE/ACM Transactions on Audio, Speech, and Language Processing*, *23*(12), 2410–2421.

Friedman, R. (2023). Tokenization in the Theory of Knowledge. *Encyclopedia, 3*(1), 380-386.

Hofmann, V., Pierrehumbert, J., y Schütze, H. (2021). Superbizarre is not superb: Derivational morphology improves BERT's interpretation of complex words. *Proceedings of the 59th Annual Meeting of the Association for Computational*

Linguistics and the 11th International Joint Conference on Natural Language Processing (Volume 1: Long Papers)*, 3594–3608.

Jabbar, H. (2024) MorphPiece: A Linguistic Tokenizer for Large Language Models. Retrieved from https://arxiv.org/html/2307.07262v2

Kudo, T., y Richardson, J. (2018). Sentencepiece: A simple and language independent subword tokenizer and detokenizer for neural text processing. *arXiv Preprint arXiv:1808. 06226*.

Lan, Z., Chen, M., Goodman, S., Gimpel, K., Sharma, P., y Soricut, R. (2019). Albert: A lite BERT for self-supervised learning of language representations. *arXiv preprint arXiv:1909.11942*.

Liu, Y., Ott, M., Goyal, N., Du, J., Joshi, M., Chen, D., … Stoyanov, V. (2019). RoBERTa: A Robustly Optimized BERT Pretraining Approach. *arXiv [Cs.CL]*. Retrieved from http://arxiv.org/abs/1907.11692

García-Sierra, Ó., Fernández-Pampillón, A, & Ortega-Martín, M. (2024). Evaluación morfológica de los vocabularios de subpalabras utilizados por los grandes modelos de lenguaje. Revista Española de Lingüística, 54(1), 103-130.

Park, K., Lee, J., Jang, S., y Jung, D. (2020). An Empirical Study of Tokenization Strategies for Various Korean NLP Tasks. *arXiv Preprint arXiv:2010. 02534*.

Radford, A., Narasimhan, K., Salimans, T., y Sutskever, I. (2018). *Improving language understanding by generative pre-training*. https://paperswithcode.com/paper/improving-language-understanding-by

Sennrich, R., Haddow, B., y Birch, A. (2015). Neural machine translation of rare words with subword units. *arXiv Preprint arXiv:1508. 07909*.

Schuster, M., y Nakajima, K. (2012). Japanese and korean voice search. *2012 IEEE International Conference on Acoustics, Speech and Signal Processing (ICASSP)*, 5149–5152. IEEE.

Song, X., Salcianu, A., Song, Y., Dopson, D., y Zhou, D. (2020). Fast *Wordpiece* tokenization. *arXiv preprint arXiv:2012.15524*.

Suárez, P. J. O., Sagot, B., y Romary, L. (2019). Asynchronous pipeline for processing huge corpora on medium to low resource infrastructures. *7th Workshop on the Challenges in the Management of Large Corpora (CMLC-7)*. Leibniz-Institut für Deutsche Sprache.

Suárez, P. J. O., Romary, L., y Sagot, B. (2020). A monolingual approach to contextualized word embeddings for mid-resource languages. *arXiv Preprint arXiv:2006. 06202*.

Van der Wouden, T. (1990). Celex: Building a multifunctional polytheoretical lexical data base. *Proceedings of BudaLex*, *88*, 363–373.

Vaswani, A., Shazeer, N., Parmar, N., Uszkoreit, J., Jones, L., Gomez, A. N., … Polosukhin, I. (2017). Attention is all you need. *Advances in Neural Information Processing Systems*, *30*.

Wu, Y., Schuster, M., Chen, Z., Le, Q. V., Norouzi, M., Macherey, W. y Dean, J. (2016). Google's neural machine translation system: Bridging the gap between human and machine translation. *arXiv Preprint arXiv:1609. 08144*.